\documentclass[12pt]{article}
\usepackage{fancyhdr}
\usepackage{url}
\usepackage{graphicx}

 {\fancyhead{}
\fancyfoot[c]{\small{\rule{17.5cm}{1pt}}}}

\begin{document}
\title{\vspace{-2.5cm}
\begin{center}
\textbf{\small{Dagstuhl Report on Conversational Search}}\\\vspace{-0.5cm} \rule{17.5cm}{1pt}
\end{center}
\vspace{1cm}\textbf{Conversational Search for Learning Technologies}}

\author{Sharon Oviatt \\
       Monash University, Melbourne, VIC, Australia\\
       \emph{sharon.oviatt@monash.edu}
       \and
       Laure Soulier \\
       Sorbonne Université, CNRS, LIP6, F-75005 Paris, France\\
       \emph{laure.soulier@lip6.fr} \\
       \date{}}

\maketitle \thispagestyle{fancy} 

\abstract{Conversational search is based  on a user-system cooperation with the objective to solve an information-seeking task. In this report, we discuss the implication of such cooperation with the learning perspective from both user and system side. We also focus  on the stimulation of learning through a key component of conversational search, namely the multimodality of communication way, and discuss the implication in terms of information retrieval. We end with a research road map describing promising research directions and perspectives.}

\section{Context and background}

\subsection{What is Learning?}
Arguably, the most important scenario for search technology is lifelong learning and education, both for students and all citizens. Human learning is a complex multidimensional activity, which includes procedural learning (e.g., activity patterns associated with cooking, sports) and knowledge-based learning (e.g., mathematics, genetics). It also includes different levels of learning, such as the ability to solve an individual math problem correctly. It also includes the development of meta-cognitive self-regulatory abilities, such as recognizing the type of problem being solved and whether one is in an error state. These latter types of awareness enable correctly regulating one’s approach to solving a problem, and recognizing when one is off track by repairing momentary errors as needed. Later stages of learning enable the generalization of learned skills or information from one context or domain to others— such as applying math problem solving to calculations in the wild (e.g., calculation of garden space, engineering calculations required for a structurally sound building).

\subsection{Human versus System Learning}
When people engage an IR system, they search for many reasons. In the process they learn a variety of things about search strategies, the location of information, and the topic about which they are searching. Search technologies also learn from and adapt to the user, their situation, their state of knowledge, and other aspects of the learning context \cite{collinsthompson2017}. Beyond adaptation, the engagement of the system impacts the search effectiveness: its pro-activity is required to anticipate user’s need, topic drift, and lower the cognitive load of users \cite{Tang2019}. For example, when someone is using a keyboard-based IR system of today, educational technologies can adapt to the person’s prior history of solving a problem correctly or not, for example by presenting a harder problem next if the last problem was solved correctly, or presenting an easier problem if it was solved incorrectly.

Based on conversational speech IR systems, it is now possible for a system to process a person’s acoustic-prosodic and linguistic input jointly, and on that basis a system can adapt to the person’s momentary state of cognitive load. The ideal state for engaging in new learning would be a moderate state of load, whereas detection of very high cognitive load might suggest that the person could benefit from taking a break for some period of time or address easier subtopics to decomplexify the search task \cite{AwadallahWPDW14}.

\section{Motivation}
\subsection{How is Learning Stimulated?}
Based on the cognitive science and learning sciences literature, it is well known that human thought is spatialized. Even when we engage in problem-solving about temporal information, we spatialize it \cite{JohnsonLaird1996}. Since conversational speech is not a spatial modality, it is advantages to combine it with at least one other spatial modality. For example, digital pen input permits handwriting diagrams and symbols that convey spatial location and relations among objects. Further, a permanent ink trace remains, which the user can think about. Tangible input like touching and manipulating objects in a virtual world also supports conveying 3D spatial information, which is especially beneficial for procedural learning (e.g., learning to drive in a simulator). Since learning is embodied and enhanced by a person’s physical activity, touch, manipulation, and handwriting can spatialize information and result in a higher level of interactivity, producing more durable and generalizable learning. When combined with conversational input for social exchange with other people, such input supports richer multimodal input.

Based on the information-seeking point of view, the understanding of users’ information need is crucial to maintain their attention and improve their satisfaction. As of now, the understanding of information need has been evaluated using relevant documents, but it implies a more complex process dealing with information need elicitation due to its formulation in natural language \cite{Aissa2018} and information synthesis \cite{Marchionini06,White2009}. There is, therefore, a crucial need to build information retrieval systems integrating human goals.

\subsection{How Can We Benefit from Multimodal IR?}
Multimodality is the preferred direction for extending conversational IR systems to provide future support for human learning. A new body of research has established that when a person can use multimodal input to engage a system, all types of thinking and reasoning are facilitated, including (1) convergent problem solving (e.g., whether a math problem is solved correctly); (2) divergent ideation (e.g., fluency of appropriate ideas when generating science hypotheses); and (3) accuracy of inferential reasoning (e.g., whether correct inferences about information are concluded or the information is overgeneralized) \cite{Oviatt2013}. It is well recognized within education that interaction with multimodal/multimedia information supports improved learning. It also is well recognized that this richer form of information enables accessibility for a wider range of diverse students (e.g., blind and hearing impaired, lower-performing, non-native speakers) \cite{Oviatt2013}.

For these and related reasons, the long-term direction of IR technologies would benefit by transitioning from conversational to multimodal systems that can substantially improve both the depth and accessibility of educational technologies. With respect to system adaptivity, when a person interacts multimodally with an IR system, the system now can collect richer contextual information about his or her level of domain expertise \cite{Oviatt2018}. When the system detects that the person is a novice in math, for example, it can adapt by presenting information in a conceptually simpler form and with fewer technical terms. In contrast, when a person is detected to be an expert, the system can adapt by upshifting to present more advanced concepts using domain-specific terminology and greater technical detail. This level of IR system adaptivity permits targeting information delivery more appropriately to a given person, which improves the likelihood that he or she will comprehend, reuse, and generalize the information in important ways. The more basic forms of system adaptivity are maintained, but also substantially expanded by the integration of more deeply human-centered models of the person and their existing knowledge of a particular content domain.

Apart from the greater sophistication of user modeling and improved system adaptivity, multimodal IR systems would benefit significantly by becoming more robust and reliable at interpreting a person’s queries to the system, compared with a speech-only conversational system \cite{Oviatt2015}. This is because fusing two or more information sources reduces recognition errors. There are both human-centered and system-centered reasons why recognition errors can be reduced or eliminated when a person interacts with a multimodal system. First, humans will formulate queries to the IR system using whichever modality they believe is least error-prone, which prevents errors. For example, they may speak a query, but switch to writing when conveying surnames or financial information involving digits. In addition, when they encounter a system error after speaking input, they can switch to another modality like writing information or even spelling a word—which leads to recovering from the error more quickly. When using a speech-only system, instead the person must re-speak information, which typically causes them to hyperarticulate. Since hyperarticulate speech departs farther from the system’s original speech training model, the result is that system errors typically increase rather than resolving successfully \cite{Oviatt2015}.

\subsection{How can user learning and system learning function cooperatively in a multimodal IR framework?}
\begin{figure}
    \centering
    \includegraphics[width=0.8\textwidth]{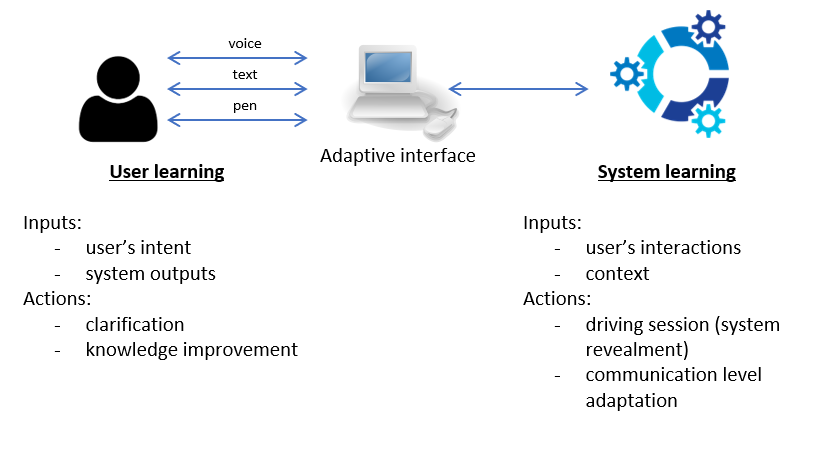}
    \caption{User Learning and System Learning in Conversational Search.}
    \label{fig}
\end{figure}

Conversational search needs to be supported by multimodal devices and algorithmic systems trading off search effectiveness and users’ satisfaction \cite{Tang2019}. Figure \ref{fig} illustrates how the user, the system, and the multi-modal interface might cooperate.
The conversation is initiated by users who formulate their information need through a modality (voice, text, pen, …). The system is expected to be proactive by fostering both 1) user revealment by eliciting the information need and 2) system revealment by suggesting what actions are available at the current state of the session \cite{Azzopardi2018}. In response, users are able to clarify their need and the span of the search session, providing them a deeper knowledge with respect to their information need. The relevant features impacting both users and system’s actions include 1) users’ intent, 2) users’ interactions, 3) system outputs, and 4) the context of the session (communication modality, spatial and temporal information…).
Several advantages of the user and system cooperation might be noticed. First, based on past interactions, the system is able to learn from right and wrong past actions. It is, therefore, more willing to target IR pieces of information that might be relevant to users. This straightforward allows reducing interactions between users and systems and lower the cognitive effort of users. Second, users being driven by increasing their knowledge acquisition experience, the system should be able to learn users’ satisfaction and therefore bolster new information in the retrieval process. Altogether, these advantages advocate for a more sophisticated and a deeper user modeling regarding both knowledge and retrieval satisfaction.

\section{Research Directions and Perspectives}
\textbf{Proposed Research and Challenges: Directions for the Community and Future PhD Topics.}
Among the key research directions and challenges to be addressed in the next 5-10 years in order to advance conversational search as a more capable learning technology are the following:
\begin{itemize}
    \item Transforming existing IR knowledge graphs into richer multi-dimensional ones that currently are used in multimodal analytic research — which supports integrating information from multiple modalities (e.g., speech, writing, touch, gaze, gesturing) and multiple levels of analyzing them (e.g., signals, activity patterns, representations).
    \item Integration of multimodal input and multimedia output processing with existing IR techniques
    \item Integration of more sophisticated user modeling with existing IR techniques, in particular ones that enable identifying the user’s current expertise level in the content domain that is the focus of their search and leveraging the span of the search session.
    \item Conversely, integrating analytics that enable the user to identify the authoritativeness of an information source (e.g., its level of expertise, its credibility or intent to deceive).
    \item Development of more advanced multimodal machine learning methods that go beyond audio-visual information processing and search.
Development of more advanced machine learning methods for extracting and representing multimodal user behavioral models.
\end{itemize}

\textbf{Broader Impact.}
The research roadmap outlined above would result in major and consequential advances, including in the following areas:
\begin{itemize}
    \item More successful IR system adaptivity for targeting user search goals.
    \item IR systems that function well based on fewer and briefer interactions between user and system.
    \item IR system that are more reliable and robust at processing user queries.
Expansion of the accessibility of IR technology to a broader population.
    \item Improved focus of IR technology on end-user goals and values, rather than commercial for-profit aims.
    \item Improvement of powerful machine learning methods for processing richer multimodal information and achieving more deeply human-centered models.
    \item Acceleration of the positive impact of lifelong learning technologies on human thinking, reasoning, and deep learning.
\end{itemize}

\textbf{Obstacles and Risks.}
\begin{itemize}
    \item Establishing and integrating more deeply human-centered multimodal behavioral models to advance IR technologies risks privacy intrusions that must be addressed in advance.
    \item Establishing successful multidisciplinary teamwork among IR, user modeling, multimodal systems, machine learning, and learning sciences experts will need to be cultivated and maintained over a lengthy period of time.
    \item Mutually adaptive systems risk unpredictability and instability of performance, and must be studied to achieve ideal functioning.
    \item New evaluation metrics will be required that substantially expand those used by IR system developers today.
\end{itemize}

\section{Acknowledgements}
This report is a summary of breaking groups during the Dagstuhl Seminar 19461 on Conversational Search. We would like to thank the Schloss Dagstuhl and the seminar organizers Avishek Anand, Lawrence Cavedon, Hideo Joho, Mark Sanderson, and Benno Stein for this week of research introspection and networking.
We also thank the ANR project SESAMS (Projet-ANR-18-CE23-0001) which supports Laure Soulier's work on this topic.

\bibliographystyle{alpha}
\bibliography{biblio}
\end{document}